\documentclass[runningheads]{llncs}

\usepackage[T1]{fontenc}
\usepackage[utf8x]{inputenc}

\usepackage{graphicx}
\usepackage[dvipsnames]{xcolor}
\usepackage{hyperref}

\usepackage{array}
\usepackage{booktabs}
\usepackage{makecell}
\usepackage{siunitx}
\usepackage{multirow}

\usepackage{siunitx}
\usepackage[misc]{ifsym}

\newcolumntype{C}[1]{>{\centering\let\newline\\\arraybackslash\hspace{0pt}}m{#1}}

\begin{document}

\title{Voxel Scene Graph for Intracranial Hemorrhage}

\author{
Antoine P. Sanner\inst{1,2} \orcidID{0000-0002-4917-9529} \textsuperscript{(\Letter)}  \and
Nils F. Grauhan\inst{2} \and
Marc A. Brockmann\inst{2} \and
Ahmed E. Othman\inst{2} \and
Anirban Mukhopadhyay\inst{1}
}
\index{Antoine Sanner}
\index{Nils Grauhan}
\index{Marc Brockmann}
\index{Ahmed Othman}
\index{Anirban Mukhopadhyay}

\authorrunning{A. Sanner et al.}
\institute{Department of Computer Science, Technical University of Darmstadt, Germany \\ 
\and
Department of Neuroradiology, University Medical Center Mainz, Germany \\
\email{antoine.sanner@gris.tu-darmstadt.de}
}

\maketitle              

\begin{abstract}

Patients with Intracranial Hemorrhage (ICH) face a potentially life-threatening condition, and patient-centered individualized treatment remains challenging due to possible clinical complications. 
Deep-Learning-based methods can efficiently analyze the routinely acquired head CTs to support the clinical decision-making. The majority of early work focuses on the detection and segmentation of ICH, but do not model the complex relations between ICH and adjacent brain structures. 
In this work, we design a tailored object detection method for ICH, which we unite with segmentation-grounded Scene Graph Generation (SGG) methods to learn a holistic representation of the clinical cerebral scene. 
To the best of our knowledge, this is the first application of SGG for 3D voxel images. 
We evaluate our method on two head-CT datasets and demonstrate that our model can recall up to 74\% of clinically relevant relations. This work lays the foundation towards SGG for 3D voxel data.
The generated Scene Graphs can already provide insights for the clinician, but are also valuable for all downstream tasks as a compact and interpretable representation.

\end{abstract}

\section{Introduction}
\label{sec:intro}

Intracranial Hemorrhage (ICH) is a potentially life-threatening emergency in which rapid recognition and decision on further treatment is crucial for patient survival \cite{Greenberg2022-wo,Hemphill2015-ke}. The 1-year mortality rate is up to 40\% in various studies \cite{An2017-lu,Pinho2019-fq} and survivors retain significant persistent functional limitations in two thirds of cases \cite{Moon2008-zq}. Patient-centered individualized treatment decisions remain clinically challenging, especially considering that patient outcome deteriorates rapidly after ICH onset \cite{Al-Shahi_Salman2018-he}. Given that CT imaging remains the main tool for diagnosis and planning, Deep Learning-based solutions can leverage this data to improve ICH patients' care, e.g. reducing the time to diagnosis.

The majority of early Deep Learning work focuses on the detection and segmentation of ICH \cite{Cho2019-az,Kuo2019-ix}, but completely ignore the related clinical complications. Indeed, hemorrhages can interact with other brain structures. Through hemorrhage expansion or involvement of the ventricular system, ICH can cause severe complications \cite{Kuo2019-ix}, which are therefore additional predictors for patient outcome \cite{Garton2016-zg}. Likewise, a bleeding-induced midline shift is a strong predictor of poor patient outcome \cite{Xu2023-ay}. However, not all bleedings are relevant, such as smaller ones that do not interact with other anatomies. As such, the true potential of automated solutions lies in modeling and understanding complex relations between ICH and adjacent brain structures to build a \textbf{clinical cerebral scene}, that can be then used for downstream tasks such as patient outcome prediction.

\begin{figure}[t]
\centering
  \includegraphics[width=.8\linewidth]{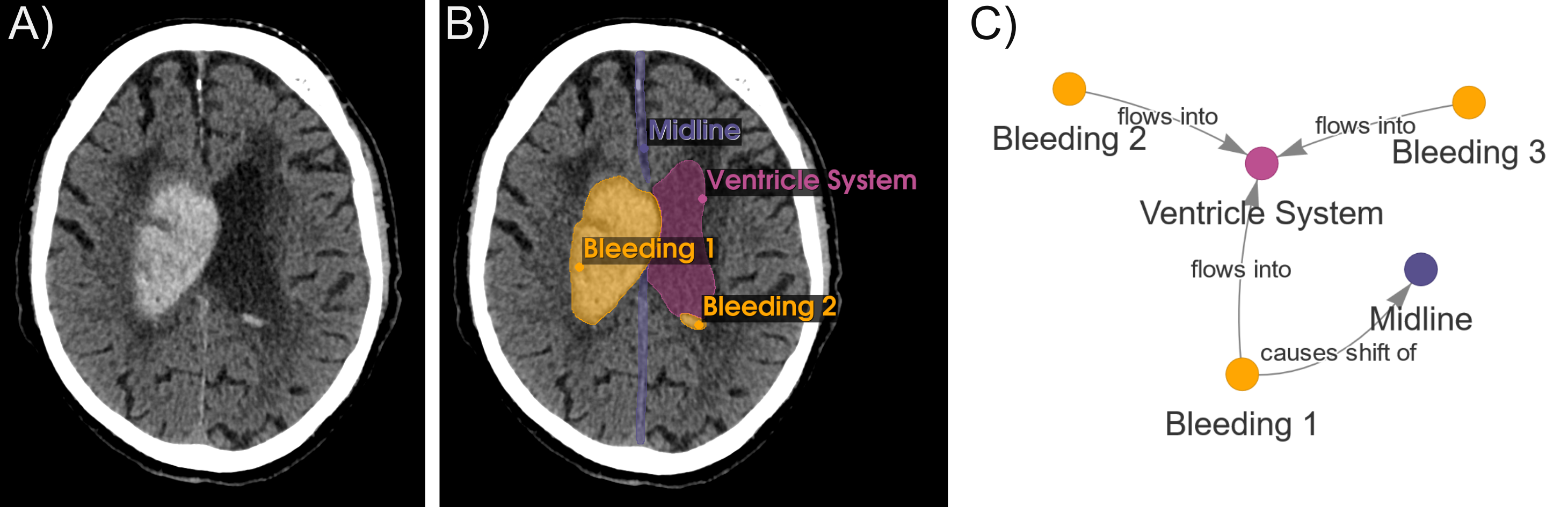}\qquad

\caption{
Example of Scene Graph for ICH. A) A slice of CT scan from an ICH patient. B) The slice with object localization. C) The associated Scene Graph. 
}
\label{fig:sg_example}
\end{figure}

Scene Graph Generation (SGG), originally introduced for natural scene-understanding \cite{zhu2022scene}, has already been applied in surgical data science \cite{zsoy2023} to model surgical procedures in the operating room. SGG techniques can likewise show their potential, when used with voxel data such as 3D CT scans. Such an application to cerebral scene for ICH is shown in Fig. \ref{fig:sg_example}. Scene Graphs offer a compact and interpretable representation, that goes beyond pure object detection and can already provide insights for the clinician.
Additionally, an advantage of Scene Graphs is built-in explainability for downstream tasks.

SGG is a two-step process, which first involves localizing relevant objects in an image. Inspired by Retina U-Net \cite{DBLP:journals/corr/abs-1811-08661}, we design a hybrid method that leverages both bounding box annotation and semantic segmentation to both detect objects robustly, and to overcome the challenges of 3D object matching (i.e. orientation-dependent aspect ratios, and significant object overlap). Then, relations between pairs of objects are predicted with SGG methods. Inspired by \textit{Neural Motifs} \cite{DBLP:journals/corr/abs-1711-06640}, and \textit{Iterative Message Passing} \cite{xu2017scene}, we design two variants for relation prediction for 3D voxel data (V-MOTIF and V-IMP). These use Recurrent Neural Networks to build a global context and from which the relation of each bleeding to adjacent brain structures is predicted.

We show that our model can recall up to 74\% of clinically relevant relations. Our contributions include:
\begin{enumerate}
    \item To the best of our knowledge, this is the first time Scene Graphs are generated from 3D voxel data. Scene Graphs can already provide insights for the clinician, independently of their added benefits for downstream tasks.
    
    \item A tailored object detection method\footnote{Code available at \url{https://github.com/MECLabTUDA/VoxelSceneGraph}} for ICH, which outperforms the state-of-the-art \textit{nnDetection} \cite{incomplete10.1007/978-3-030-87240-3_51}, and which we unite with our voxel-adapted SGG methods to learn a holistic representation of the clinical cerebral scene.

\end{enumerate}

\section{Methodology}

\begin{figure}
    \centering
        \centering
        \includegraphics[width=\textwidth]{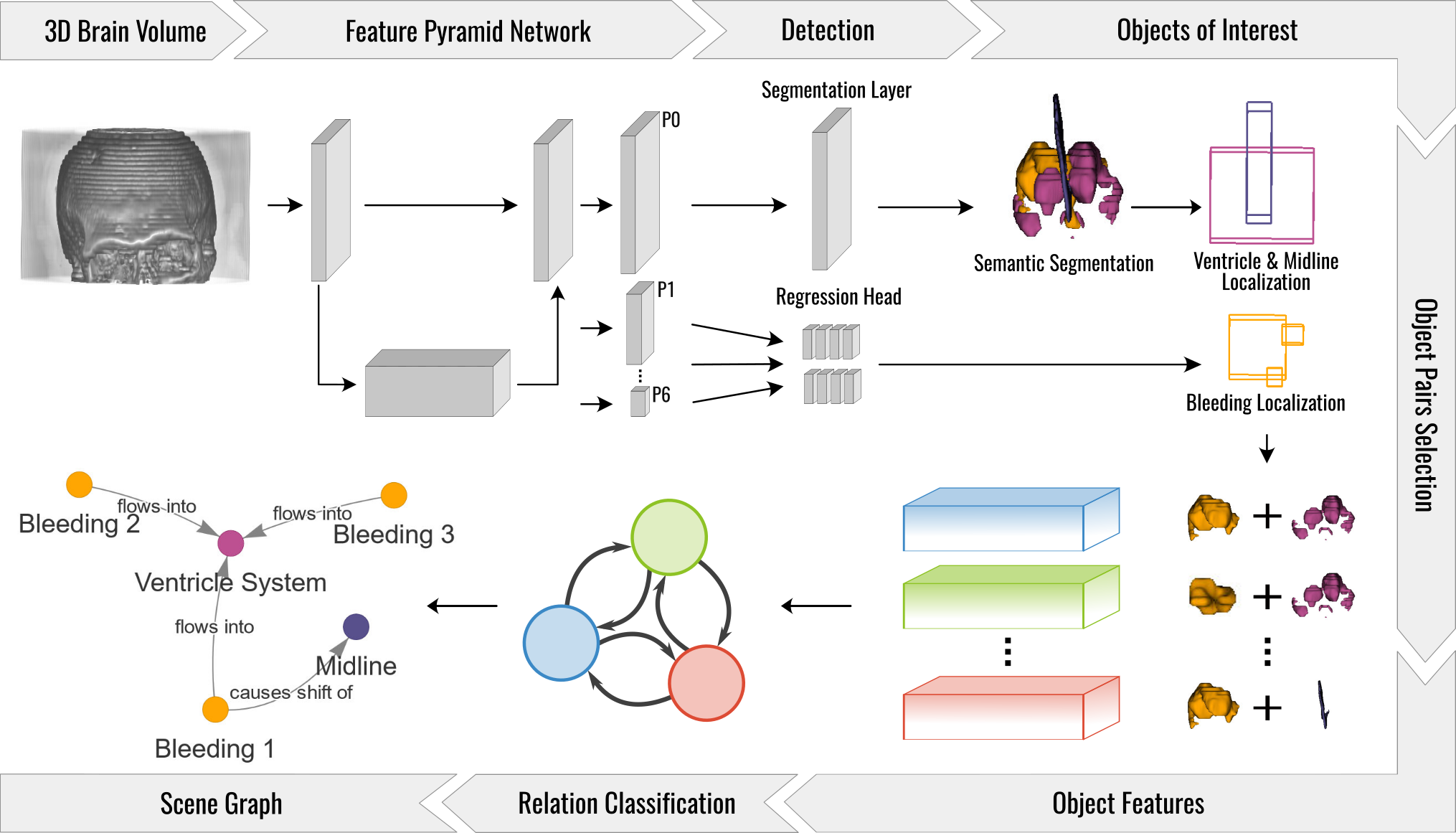}
    \caption{Overview of our two-stage method for Scene Graph Generation. Objects are first detected using a hybrid detector / segmentation model. The relations are then predicted using both bounding box and segmentation mask information.}
    \label{fig:meth}
\end{figure}

In this section, we give insights into how we structured Scene Graphs for ICH. We then introduce our two-stage method for SGG as visualized in Fig. \ref{fig:meth}.

\label{sec:data}
\noindent\textbf{Scene Graph for ICH:} The first and most critical step is to define which structures are relevant, as well as a set of relations that can model possible sources of complications. We select the following interactions for our application based on possible clinical complications: 1) midline shifts, 2) blood flow to the ventricle system, and 3) swelling induced asymmetry of the ventricle system. As such, we have \textbf{three classes of objects} ("Bleeding", "Ventricle System" and "Midline"), and \textbf{three bleeding-induced classes of relations}. Additionally, we expect all patients to have exactly one midline. We model the ventricle system as a single object even when it is only sparsely visible, since the presence of blood can have severe complications independently of the location within the system.

\noindent\textbf{Object Detection:} Localizing individual bleeding requires a multiscale approach, as they can have vastly different size, shape, and position. The \textbf{3D Retina-UNet} coupled with a Feature Pyramid Network has already proved its usefulness for 3D medical imaging \cite{DBLP:journals/corr/abs-1811-08661}. In particular, this architecture is very flexible and can fully leverage multiscale features. This is crucial as the volume of the bleeding can range from only $\SI{0.1}{\centi\meter}^3$ to more than $\SI{100}{\centi\meter}^3$. 
In opposition, the midline and the ventricle system only appear at a single scale, but pose their own set of challenges. 
Indeed, the 3D aspect ratio of the midline will heavily depend on the head's orientation, especially in the axial plane. This both can cause issues with anchor matching and bounding box regression.
In contrast, depending on the swelling or the presence of blood, the ventricle system may appear as multiple fragmented objects. These would need to be detected individually, as there would be a large overlap between the bleeding with the system and the system itself, rendering an accurate object matching impossible. For these reasons, we decide to leverage Retina-UNet's segmentation capabilities by detecting both anatomies from the predicted semantic segmentation. This simple but elegant solution, can solve the previously mentioned challenged, while also ensuring that only one object is predicted per class per image.

\noindent\textbf{Relation Prediction:} Once objects are localized in the image, SGG methods focus on building a global context from individual object features (as defined by \cite{Tang2020}). Inspired by \textit{Neural Motifs} \cite{DBLP:journals/corr/abs-1711-06640}, and \textit{Iterative Message Passing} \cite{xu2017scene}, we design two variants for relation prediction for 3D voxel data (\textbf{V-MOTIF} and \textbf{V-IMP}). 

\textbf{V-MOTIF} uses bidirectional Long Short-term Memory Networks and iterates over detected objects. We choose to order these objects from top-to-bottom, as related objects in the cerebral scene are very likely to be at the same depth. Sorting by size, may be another viable solution, but may create a bias towards certain relation classes.
In comparison, \textbf{V-IMP} builds a primal object graph, and its dual edge graph, then combines Gated Recurrent Units (GRU) with message passing iteratively to allow for information flow. Since each GRU receives multiple messages at each iteration, learnable weights are used to pool messages while only keeping the relevant information.
Both models finally predict relations from the enriched features.
Usually, these are also used to refine the classification of each object. 
Given the object matching challenges discussed previously, we decide to only use SGG methods for relation prediction. 

Finally, \textbf{segmentation-grounding} \cite{Khandelwal2021} involves computing the object features with the object segmentation, rather than with its bounding box, to leverage the finer localization. 
Since only a semantic segmentation in computed in the first step, we use it as a proxy by cropping it to the object's bounds and binarizing it based on the object's predicted class. 

\section{Experiments}
\label{sec:exp}

In this section, we introduce the base images used and their annotation process. We then describe our evaluation setup for the different stages of our method.\\

\begin{figure}
    \centering
        \includegraphics[width=.8\textwidth]{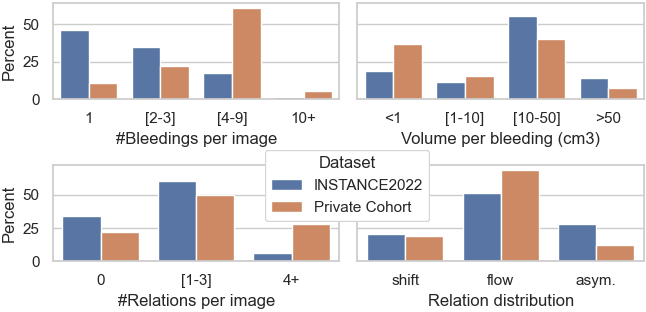} 
    \caption{Distribution of the number of bleedings per image (upper left) and volume per bleeding (upper right) for each dataset. Distribution of the number of relations per image (bottom left) and the distribution of relations (bottom right) for each dataset.}
    \label{fig:data_stats}
\end{figure}

\noindent\textbf{Source Images:} For this study, we source two datasets. The first one is the publicly available INSTANCE2022 challenge dataset \cite{li2023stateoftheart}. 
It contains 130 non-contrast head CTs of patients diagnosed with ICH. 10 images with significant streak artifacts were discarded, as they were deemed out-of-distribution. 
Additionally, we select a private cohort for the purpose of external validation. It is constituted of 18 non-contrast head CTs of patients diagnosed with ICH. We choose more severe cases (caused by trauma) compared to the first dataset.
\textbf{These cases are clinically more challenging}, e.g. some patients were already operated on, and we want to test the robustness of our method against distribution shift (see Fig. \ref{fig:data_stats}).

\noindent\textbf{Data Annotation:} The patient's head position is first manually harmonized in all images. A senior neuroradiologist then segments a label map for each image using \textit{3D Slicer} \cite{Fedorov2012-qh}, which takes 5 to 30 min per image. 3D bounding box information is automatically extracted from the label maps by computing the bounds of each segmented object.
Currently, one of the main obstacles of Scene Graph application using voxel data is the lack of tools for relation annotation. In particular, the annotator needs to be able to easily scroll through the 3D volume to select individual objects as being part of a relation. We developed an internal tool specifically to this end. We plan to make it open-source in the near future. On average, annotating relations takes under 2 min per image.

\noindent\textbf{Model Training:} For bounding box detection, we split the official training images from the INSTANCE2022 dataset \cite{li2023stateoftheart} for training and validation in an 80/20 fashion. The official validation cases are used for in-distribution testing. The private cohort is fully used for external validation. However, not all images contain relations. As such, we only keep cases with relations for training the relation detector, the splits remaining otherwise unchanged. Configuration files with hyperparameters and  data splits will be made available with the code.

\noindent\textbf{Evaluation Metrics:} For bleeding detection, we use the well established metrics for object detection, i.e. Average Recall (AR) and mean Average Precision (mAP). We choose a localization threshold of 30\% Intersection Over Union (IoU), as it reflects a balance between the clinical need for coarse localization and producing sensible bounding boxes for larger bleeding \cite{incomplete10.1007/978-3-030-87240-3_51}. 
Additionally, since no relation can get predicted when at least one object has not been detected, we also give an upper bound of how many relations can be recalled given the object detected in the first stage. 
For relation prediction, we compute the metrics based on the top-K predictions per image, with again a 30\% IoU threshold for object localization. We use the Recall@K (R@K), mean Recall@K (mR@K), and mean Average Precision@K (mAP@K). Given that an image has on average overall 2 relations (up to 7), we compute these metrics for $K=8$. Our method is evaluated for both \textbf{Predicate Classification} and \textbf{Scene Graph Generation} tasks \cite{Tang2020}, i.e. respectively predicting relations from ground truth and predicted object localization. \textbf{The results are averaged over 5 random seeds}.

\section{Results}
\label{sec:res}

In this section, we evaluate our two-stage pipeline, first for object detection and then for relation prediction (Predicate Classification and Scene Graph Generation). Results for ablations of individual method components can be found in the Supplementary Material.

\subsection{Object Detection}

\begin{figure}
    \centering
        \centering
        \includegraphics[width=\textwidth]{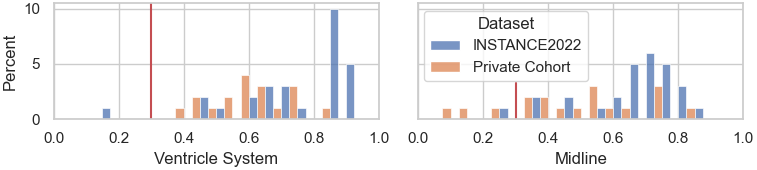} 
    \caption{IoU distribution of localized ventricle systems (left) and midlines (right) for each dataset. An anatomy is correctly detected if its bounding box IoU with the ground truth annotation is above 30\% (red line).}
    \label{fig:seg_detec}
\end{figure}

\noindent\textbf{Ventricle System \& Midline Detection:} We first evaluate the performance for detecting the ventricle system and the midline, as these structures are localized using the predicted semantic segmentation. As shown in Fig. \ref{fig:seg_detec}, our solution can detect them robustly and precisely, with a detection rate of 96.4\% for both structures in the INSTANCE2022 dataset. For the private cohort, it still has a detection rate of 100\% and 83.3\% respectively for the ventricle system and the midline.\\

\begin{table}
\centering
\begin{tabular}{l C{1.2cm}C{1.2cm}C{1.2cm}C{1.2cm} C{1.2cm} C{1.2cm} C{1.2cm} C{1.2cm}}
\toprule
& \multicolumn{4}{c}{Bleeding Detection} & \multicolumn{4}{c}{Relation Detection Upper Bound}\\
\cmidrule(lr){2-5}\cmidrule(lr){6-9}
& \multicolumn{2}{c}{INSTANCE2022} & \multicolumn{2}{c}{Private Cohort}& \multicolumn{2}{c}{INSTANCE2022} & \multicolumn{2}{c}{Private Cohort}\\
\cmidrule(lr){2-3}\cmidrule(lr){4-5}\cmidrule(lr){6-7}\cmidrule(lr){8-9}
Method &  AR$_{30}\uparrow$  & AP$_{30}\uparrow$ & AR$_{30}\uparrow$ & AP$_{30}\uparrow$&  R@8$\uparrow$ & mR@8$\uparrow$ & R@8$\uparrow$ & mR@8$\uparrow$\\
\midrule
nnDetection & 63.1 & 54.9 & 43.9 & 26.3 & 85.0 & 86.4 & 54.6 & 59.0\\
Ours & \bf 84.7 & \bf 73.1 & \bf 55.4 & \bf 43.5& \bf 96.3 & \bf 97.0 & \bf 89.1 & \bf 86.9\\
\bottomrule
\end{tabular}
\caption{Left: Bleeding detection at 30\% IoU using our method, and \textit{nnDetection} \cite{incomplete10.1007/978-3-030-87240-3_51} (Model RetinaUNetV001). Right: Upper bounds for relation prediction given the objects that are detected. Metrics for \textit{nnDetection} assume perfect localization of both ventricles system and midline.
}
\label{res:detec}
\end{table}

\noindent\textbf{Bleeding Detection:} We then evaluate the detection performance for bleeding separately from the previously mentioned structures. In Table \ref{res:detec}, we compare our method to state-of-the-art nnDetection \cite{incomplete10.1007/978-3-030-87240-3_51}. Our method significantly outperforms it on both datasets, showing that careful method parametrization can outperform AutoML for this task. 
The private cohort contains proportionally many smaller bleedings, which are not present in the training data and predominantly have no associated relation. In this regard, \textbf{our method still robustly detects relevant bleedings in the private cohort}.


\subsection{Scene Graph Prediction}

\noindent\textbf{Predicate Classification:} We evaluate our \textit{Voxel Neural Motifs} (V-MOTIF) and \textit{Voxel Iterative Message Passing} (V-IMP) models for relation prediction, with and without segmentation grounding and report the results in Table \ref{res:pred_cls}. 
Given that datasets for natural scene-understanding \cite{Chacra2022} or even surgical data science  \cite{zsoy2022} comprise thousands of scenes, one could assume that a large data quantity is required for training SGG methods. These results show that \textbf{under 200 scenes are enough to generalize across medical centers} for the ICH cerebral scene.

\begin{table}
\centering

\begin{tabular}{ l l C{1.45cm}C{1.45cm}C{1.45cm}C{1.45cm}C{1.45cm} C{1.45cm}}
\toprule
&& \multicolumn{3}{c}{INSTANCE2022} & \multicolumn{3}{c}{Private Cohort}\\
\cmidrule(lr){3-5}\cmidrule(lr){6-8}

Model & Method &  R@8$\uparrow$ & mR@8$\uparrow$ & mAP@8$\uparrow$ & R@8$\uparrow$ & mR@8$\uparrow$ & mAP@8$\uparrow$\\
\midrule
V-MOTIF & Base & 70.1±4.2 & 75.3±3.8 & 55.5±2.9 & \bf 61.1±4.7 & 45.9±3.9 & \bf 44.8±8.1\\
V-MOTIF & Seg-G & \bf 73.9±1.8 & \bf 77.0±1.3 & 54.2±3.9 & 60.3±9.9 & \bf 47.1±7.7 & 43.9±4.6\\
V-IMP & Base & 66.1±7.3 & 70.7±5.1 & 52.4±4.1 & 52.7±3.0 & 45.8±5.4 & 36.4±5.6\\
V-IMP & Seg-G & 68.8±7.9 & 72.5±5.9 & \bf 56.1±6.1 & 47.6±9.4 & 38.8±4.8 & 41.8±6.8\\
\bottomrule

\end{tabular}
\caption{
Results for the \textbf{Predicate Classification} task averaged over 5 runs. The metrics are reported with segmentation-grounding (Seg-G) and without (Base).
}
\label{res:pred_cls}
\end{table}

\begin{figure}
\centering
  \includegraphics[width=\linewidth]{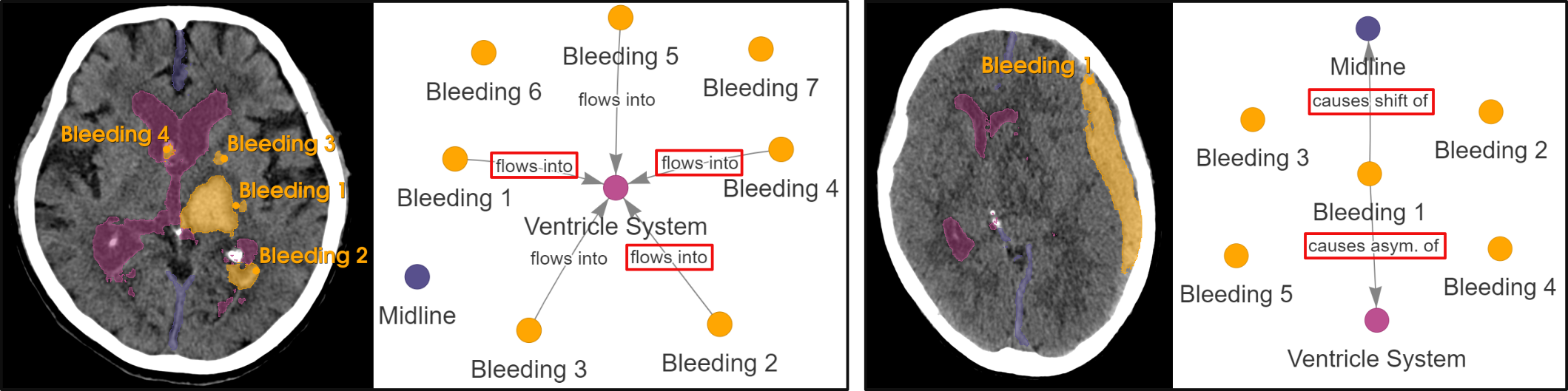}\qquad

\caption{Predicted segmentation and top-5 predictions of segmentation-grounded V-IMP on the test set of INSTANCE2022 (left) and from the private cohort (right). (Left) The model identifies that bleedings are flowing in the ventricle system, which may require a surgical intervention. (Right) A strong midline shift to the left is detected and attributed to the large subdural bleeding on the right side. Also note the asymmetry of the ventricle system. Both indicate an increased intracranial pressure, which may also require a different type of surgery than (left).}

\label{fig:qual}
\end{figure}

\noindent\textbf{Scene Graph Generation:} This experiment reflects the true performance of the entire prediction pipeline (Table \ref{res:sg_gen}). Qualitative results are shown in \linebreak Fig. \ref{fig:qual}. Overall, the models recall relevant relations, with clinically interpretable information for decision support in treatment planning. However, this experiment shows that our complete method already offers a satisfying performance for in-distribution data.  

\begin{table}[t]
\centering
\begin{tabular}{ l l C{1.45cm}C{1.45cm}C{1.45cm}C{1.45cm}C{1.45cm} C{1.45cm}}
\toprule
&& \multicolumn{3}{c}{INSTANCE2022} & \multicolumn{3}{c}{Private Cohort}\\
\cmidrule(lr){3-5}\cmidrule(lr){6-8}
Model & Method &  R@8$\uparrow$ & mR@8$\uparrow$ & mAP@8$\uparrow$ & R@8$\uparrow$ & mR@8$\uparrow$ & mAP@8$\uparrow$\\
\midrule
V-MOTIF & Base & 66.4±4.2 & 70.1±4.0 & 39.8±1.6 & 25.1±2.5 & 23.4±2.4 & 21.4±1.0\\
V-MOTIF & Seg-G & 64.2±8.9 & 71.4±5.1 & \bf 40.6±4.9 & 19.8±6.4 & 22.1±6.9 & 21.9±7.6\\
V-IMP & Base & 71.2±2.6 & 75.6±1.6 & 37.4±3.6 & \bf 30.0±2.2 & \bf 25.6±3.9 & \bf 22.3±3.4\\
V-IMP & Seg-G & \bf 74.0±6.1 & \bf 78.1±4.6 & 34.0±2.8 & 22.7±5.0 & 22.0±4.6 & 16.5±3.1\\
\bottomrule
\end{tabular}
\caption{
Results for the \textbf{Scene Graph Generation} task averaged over 5 runs. The metrics are reported with segmentation-grounding (Seg-G) and without (Base).
}
\label{res:sg_gen}
\end{table}

\section{Conclusion}

The clinical scene for Intracranial Hemorrhage (ICH) goes beyond detecting bleedings, as these can interact with adjacent brain structures, and potentially cause severe complications.
We introduce a method for structured representation learning and tailored for ICH. To the best of our knowledge, this is the first application of Scene Graph Generation (SGG) for 3D voxel data. 
We evaluate our method on two datasets and we demonstrate that our model can recall up to 74\% of clinically relevant relations.
This work paves a path towards SGG for 3D voxel images and modeling holistic anatomical scenes.
The generated Scene Graphs can be used as a compact and interpretable representation for all downstream task (e.g. patient outcome prediction) and offer a strong decision support for patient-centered treatment.
\section{Compliance with Ethical Standards}

This study was performed in line with the principles of the Declaration of Helsinki. The retrospective evaluation of imaging data from the University Medical Center Mainz was approved by the local ethics boards (Project 2021-15948-retrospektiv). Ethical approval was not required, as confirmed by the license attached with the open access data.

\begin{credits}
\subsubsection{\ackname} Antoine P. Sanner and Nils F. Grauhan equally contribute to this paper.

\subsubsection{\discintname}
The authors have no competing interests to declare that are relevant to the content of this article.
\end{credits}

%
%
%

\bibliographystyle{splncs04}
\bibliography{Paper-0751}

\begin{thebibliography}{10}
\providecommand{\url}[1]{\texttt{#1}}
\providecommand{\urlprefix}{URL }
\providecommand{\doi}[1]{https://doi.org/#1}

\bibitem{Al-Shahi_Salman2018-he}
Al-Shahi~Salman, R., Frantzias, J., Lee, R.J., Lyden, P.D., Battey, T.W.K., Ayres, A.M., Goldstein, J.N., Mayer, S.A., Steiner, T., Wang, X., Arima, H., Hasegawa, H., Oishi, M., Godoy, D.A., Masotti, L., Dowlatshahi, D., Rodriguez-Luna, D., Molina, C.A., Jang, D.K., Davalos, A., Castillo, J., Yao, X., Claassen, J., Volbers, B., Kazui, S., Okada, Y., Fujimoto, S., Toyoda, K., Li, Q., Khoury, J., Delgado, P., Sab{\'\i}n, J.{\'A}., Hern{\'a}ndez-Guillamon, M., Prats-S{\'a}nchez, L., Cai, C., Kate, M.P., McCourt, R., Venkatasubramanian, C., Diringer, M.N., Ikeda, Y., Worthmann, H., Ziai, W.C., d'Esterre, C.D., Aviv, R.I., Raab, P., Murai, Y., Zazulia, A.R., Butcher, K.S., Seyedsaadat, S.M., Grotta, J.C., Mart{\'\i}-F{\`a}bregas, J., Montaner, J., Broderick, J., Yamamoto, H., Staykov, D., Connolly, E.S., Selim, M., Leira, R., Moon, B.H., Demchuk, A.M., Di~Napoli, M., Fujii, Y., Anderson, C.S., Rosand, J., {VISTA-ICH Collaboration}, {ICH Growth Individual Patient Data Meta-analysis Collaborators}: Absolute risk and
  predictors of the growth of acute spontaneous intracerebral haemorrhage: a systematic review and meta-analysis of individual patient data. Lancet Neurol.  \textbf{17}(10),  885--894 (Oct 2018)

\bibitem{An2017-lu}
An, S.J., Kim, T.J., Yoon, B.W.: Epidemiology, risk factors, and clinical features of intracerebral hemorrhage: An update. J. Stroke  \textbf{19}(1),  3--10 (Jan 2017)

\bibitem{incomplete10.1007/978-3-030-87240-3_51}
Baumgartner, M., J{\"a}ger, P.F., Isensee, F., Maier-Hein, K.H.: nndetection: A self-configuring method for medical object detection. In: de~Bruijne, M., Cattin, P.C., Cotin, S., Padoy, N., Speidel, S., Zheng, Y., Essert, C. (eds.) Medical Image Computing and Computer Assisted Intervention -- MICCAI 2021. pp. 530--539. Springer International Publishing, Cham (2021)

\bibitem{Chacra2022}
Chacra, D.A., Zelek, J.: The topology and language of relationships in the visual genome dataset. In: 2022 IEEE/CVF Conference on Computer Vision and Pattern Recognition Workshops (CVPRW). IEEE (Jun 2022). \doi{10.1109/cvprw56347.2022.00533}

\bibitem{Cho2019-az}
Cho, J., Park, K.S., Karki, M., Lee, E., Ko, S., Kim, J.K., Lee, D., Choe, J., Son, J., Kim, M., Lee, S., Lee, J., Yoon, C., Park, S.: Improving sensitivity on identification and delineation of intracranial hemorrhage lesion using cascaded deep learning models. J. Digit. Imaging  \textbf{32}(3),  450--461 (Jun 2019)

\bibitem{Fedorov2012-qh}
Fedorov, A., Beichel, R., Kalpathy-Cramer, J., Finet, J., Fillion-Robin, J.C., Pujol, S., Bauer, C., Jennings, D., Fennessy, F., Sonka, M., Buatti, J., Aylward, S., Miller, J.V., Pieper, S., Kikinis, R.: {3D} slicer as an image computing platform for the quantitative imaging network. Magn. Reson. Imaging  \textbf{30}(9),  1323--1341 (Nov 2012)

\bibitem{Garton2016-zg}
Garton, T., Keep, R.F., Wilkinson, D.A., Strahle, J.M., Hua, Y., Garton, H.J.L., Xi, G.: Intraventricular hemorrhage: The role of blood components in secondary injury and hydrocephalus. Transl. Stroke Res.  \textbf{7}(6),  447--451 (Dec 2016)

\bibitem{Greenberg2022-wo}
Greenberg, S.M., Ziai, W.C., Cordonnier, C., Dowlatshahi, D., Francis, B., Goldstein, J.N., Hemphill, 3rd, J.C., Johnson, R., Keigher, K.M., Mack, W.J., Mocco, J., Newton, E.J., Ruff, I.M., Sansing, L.H., Schulman, S., Selim, M.H., Sheth, K.N., Sprigg, N., Sunnerhagen, K.S., {American Heart Association/American Stroke Association}: 2022 guideline for the management of patients with spontaneous intracerebral hemorrhage: A guideline from the american heart association/american stroke association. Stroke  \textbf{53}(7),  e282--e361 (Jul 2022)

\bibitem{Hemphill2015-ke}
Hemphill, 3rd, J.C., Greenberg, S.M., Anderson, C.S., Becker, K., Bendok, B.R., Cushman, M., Fung, G.L., Goldstein, J.N., Macdonald, R.L., Mitchell, P.H., Scott, P.A., Selim, M.H., Woo, D., {American Heart Association Stroke Council}, {Council on Cardiovascular and Stroke Nursing}, {Council on Clinical Cardiology}: Guidelines for the management of spontaneous intracerebral hemorrhage: A guideline for healthcare professionals from the american heart {Association/American} stroke association. Stroke  \textbf{46}(7),  2032--2060 (May 2015)

\bibitem{DBLP:journals/corr/abs-1811-08661}
Jaeger, P.F., Kohl, S.A.A., Bickelhaupt, S., Isensee, F., Kuder, T.A., Schlemmer, H., Maier{-}Hein, K.H.: Retina u-net: Embarrassingly simple exploitation of segmentation supervision for medical object detection. CoRR  \textbf{abs/1811.08661} (2018)

\bibitem{Khandelwal2021}
Khandelwal, S., Suhail, M., Sigal, L.: Segmentation-grounded scene graph generation. In: 2021 IEEE/CVF International Conference on Computer Vision (ICCV). IEEE (Oct 2021). \doi{10.1109/iccv48922.2021.01558}

\bibitem{Kuo2019-ix}
Kuo, W., H{\"a}ne, C., Mukherjee, P., Malik, J., Yuh, E.L.: Expert-level detection of acute intracranial hemorrhage on head computed tomography using deep learning. Proc. Natl. Acad. Sci. U. S. A.  \textbf{116}(45),  22737--22745 (Nov 2019)

\bibitem{li2023stateoftheart}
Li, X., Luo, G., Wang, K., Wang, H., Liu, J., Liang, X., Jiang, J., Song, Z., Zheng, C., Chi, H., Xu, M., He, Y., Ma, X., Guo, J., Liu, Y., Li, C., Chen, Z., Siddiquee, M.M.R., Myronenko, A., Sanner, A.P., Mukhopadhyay, A., Othman, A.E., Zhao, X., Liu, W., Zhang, J., Ma, X., Liu, Q., MacIntosh, B.J., Liang, W., Mazher, M., Qayyum, A., Abramova, V., Lladó, X., Li, S.: The state-of-the-art 3d anisotropic intracranial hemorrhage segmentation on non-contrast head ct: The instance challenge (2023)

\bibitem{Moon2008-zq}
Moon, J.S., Janjua, N., Ahmed, S., Kirmani, J.F., Harris-Lane, P., Jacob, M., Ezzeddine, M.A., Qureshi, A.I.: Prehospital neurologic deterioration in patients with intracerebral hemorrhage. Crit. Care Med.  \textbf{36}(1),  172--175 (Jan 2008)

\bibitem{zsoy2023}
{\"O}zsoy, E., Czempiel, T., Holm, F., Pellegrini, C., Navab, N.: LABRAD-OR: Lightweight Memory Scene Graphs for Accurate Bimodal Reasoning in Dynamic Operating Rooms, p. 302–311. Springer Nature Switzerland (2023). \doi{10.1007/978-3-031-43996-4_29}

\bibitem{zsoy2022}
\"{O}zsoy, E., \"{O}rnek, E.P., Eck, U., Czempiel, T., Tombari, F., Navab, N.: 4D-OR: Semantic Scene Graphs for OR Domain Modeling, p. 475–485. Springer Nature Switzerland (2022). \doi{10.1007/978-3-031-16449-1_45}

\bibitem{Pinho2019-fq}
Pinho, J., Costa, A.S., Ara{\'u}jo, J.M., Amorim, J.M., Ferreira, C.: Intracerebral hemorrhage outcome: A comprehensive update. J. Neurol. Sci.  \textbf{398},  54--66 (Mar 2019)

\bibitem{Tang2020}
Tang, K., Niu, Y., Huang, J., Shi, J., Zhang, H.: Unbiased scene graph generation from biased training. In: 2020 IEEE/CVF Conference on Computer Vision and Pattern Recognition (CVPR). IEEE (Jun 2020). \doi{10.1109/cvpr42600.2020.00377}

\bibitem{xu2017scene}
Xu, D., Zhu, Y., Choy, C.B., Fei-Fei, L.: Scene graph generation by iterative message passing (2017)

\bibitem{Xu2023-ay}
Xu, X.M., Zhang, H., Meng, R.L.: Cranial midline shift is a predictor of the clinical prognosis of acute cerebral infarction patients undergoing emergency endovascular treatment. Sci. Rep.  \textbf{13}(1) (Nov 2023)

\bibitem{DBLP:journals/corr/abs-1711-06640}
Zellers, R., Yatskar, M., Thomson, S., Choi, Y.: Neural motifs: Scene graph parsing with global context. CoRR  \textbf{abs/1711.06640} (2017)

\bibitem{zhu2022scene}
Zhu, G., Zhang, L., Jiang, Y., Dang, Y., Hou, H., Shen, P., Feng, M., Zhao, X., Miao, Q., Shah, S.A.A., Bennamoun, M.: Scene graph generation: A comprehensive survey (2022)

\end{thebibliography}

\end{document}